\documentclass[letterpaper, 10 pt, conference]{ieeeconf}
\IEEEoverridecommandlockouts
\usepackage{amsmath,amssymb}
\usepackage{graphicx}
\usepackage{algorithm}
\usepackage{algorithmic}
\usepackage{hyperref}
\usepackage{booktabs}
\usepackage{microtype}
\usepackage[utf8]{inputenc}

\usepackage{cite}
\usepackage{booktabs}
\usepackage{colortbl}
\usepackage{xcolor}
\usepackage{gensymb}
\usepackage{amsmath}
\usepackage{makecell}

\title{Assistance Without Interruption: A Benchmark and LLM-based Framework for Non-Intrusive Human–Robot Assistance}

\author{Yuedi Zhang, Shuanghao Bai, Wanqi Zhou, Haoran Zhang, Qi Zhang, Zhirong Luan, Badong Chen 
\thanks{Corresponding author: Badong Chen.}
\thanks{Yuedi Zhang, Shuanghao Bai, Wanqi Zhou, Haoran Zhang, Qi Zhang, and Badong Chen are with the Institute of Artificial Intelligence and Robotics, Xi’an Jiaotong University, Xi'an, China (email: zyd993@stu.xjtu.edu.cn; baishuanghao@stu.xjtu.edu.cn; zwq785915792@stu.xjtu.edu.cn; chenbd@mail.xjtu.edu.cn).} 
\thanks{Zhirong Luan is with the School of Electrical Engineering, Xi’an University of Technology, Xi'an, China (email: luanzhirong@xaut.edu.cn).}
}

\date{}

\begin{document}
\maketitle

\begin{abstract}\label{sec:abs}
Human–robot interaction (HRI) has long studied how agents and people coordinate to achieve shared goals. In this work, we formalize and benchmark the non-intrusive assistance as an independent paradigm of HRI, where a robot proactively supports a human’s ongoing multi-step activities while strictly avoiding interruptions. 
Unlike conventional HRI tasks that rely on direct commands, explicit negotiation, or proactive interventions based on user habits and history, our task treats the human’s plan as the primary process and formulates assistance as a joint decision over when to act and what to do. 
To systematically evaluate this problem, we establish a simulation benchmark, NIABench, along with new metrics tailored to the non-intrusive assistance task. 
We further propose a hybrid architecture that integrates an LLM with a scoring model. The scoring model first applies semantic retrieval to prune large candidate action sets, and then a ranker evaluates human-step and robot-action pairs, enabling reasoning over timing and cross-step dependencies. 
Comprehensive experiments on both NIABench and real-world scenarios demonstrate that our method achieves proactive, non-intrusive assistance that reduces human effort while preserving task effectiveness. 
Our benchmark and method are available at \url{https://renytek13.github.io/assistance-without-interruption}.
\end{abstract}

\section{Introduction}
\label{sec:intro}

Human–robot interaction (HRI) is broadly defined as the science and engineering of studying, designing, and evaluating interactions between humans and robots, and has long examined how agents and people coordinate to achieve shared goals~\cite{goodrich2008human}. Most existing approaches~\cite{tellex2011, hemachandra2015learning, misra2016tell, huang2022language, thumm2025text2interaction, grannen2025vocal} rely on explicit human natural language instructions as the primary medium of interaction, directly commanding or controlling the robot (e.g., “bring me a glass of water” or “peel an apple”), a paradigm often referred to as reactive assistance. Such purely reactive robotic responses, however, may result in inefficiency and can increase human workload or reduce the naturalness of interaction.

Proactivity has been identified as a central direction for future HRI~\cite{li2023proactive}. Several studies have enabled robots to proactively observe and predict human intent and actions to guide subsequent behaviors~\cite{koppula2015anticipating, maeda2017phase, nikolaidis2017human, jain2018recursive, landi2019prediction, belardinelli2024gaze, hoffman2024inferring}. While these approaches enhance coordination and responsiveness, they remain largely confined to short-term predictions tied to observed human behaviors. Moving further, proactive assistance explicitly shifts initiative to the robot~\cite{wu2016watch, shafti2019gaze, patel2023proactive, mascaro2023hoi4abot}. For example, domestic robots can remind users of forgotten actions~\cite{wu2016watch}, leverage gaze cues to assist reaching and grasping~\cite{shafti2019gaze}, or anticipate future tasks from object interactions to act in advance~\cite{mascaro2023hoi4abot}. Although these studies mark significant progress toward proactive and context-aware support, they often intervene directly in human tasks, thereby introducing additional cognitive load and resumption costs~\cite{lee2018interruption, hirsch2024examining}.

\begin{figure}[t]
  \centering
  \includegraphics[width=0.45\textwidth]{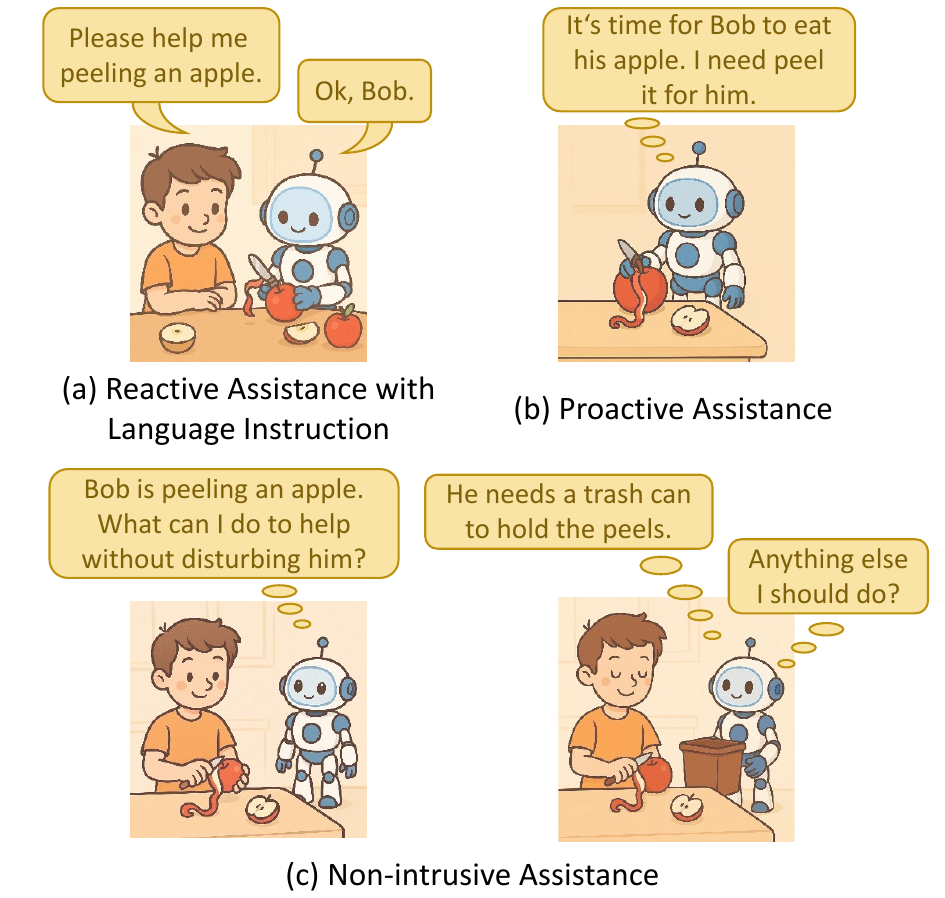}
  \caption{Modes of interaction in HRI include (a) reactive execution of commands, typically through language instructions, and (b) proactive intervention based on user habits and history. 
  In contrast to these tasks, we introduce (c) non-intrusive assistance, where the robot supports human tasks without causing disruption. 
  }
  \label{fig:intro}
\end{figure}

Human daily activities often unfold as long, continuous sequences of small, interleaved actions. As illustrated in Figure~\ref{fig:intro}, consider a kitchen scenario where a person prepares an apple to eat: washing it, setting it down, minutes later reaching for a knife to slice it, and finally disposing of the peels. In such settings, assistance that arrives quietly and at the right time, such as placing a plate or bringing a trash bin, can substantially reduce human effort while preserving the natural flow of activity. 
Motivated by this observation, we formalize Non-intrusive Assistance (NIA), an independent HRI assistance paradigm that is fundamentally different from reactive and proactive assistance, which focuses on proactively helping humans complete long-horizon activities without disrupting their ongoing progress. 
While some existing proactive assistance studies have explored non-intrusive design tendencies~\cite{schmid2007proactive, mollaret2016multi}, these works still fail to regard the human task as an unalterable primary process and absolute non-interruption. 
Distinct from these, NIA takes the human’s task plan as the unmodifiable primary process, where the robot acts as a silent assistance agent without explicit human requests, no robot-induced interruption, and zero additional cognitive load for humans, focusing on proactively helping humans complete ongoing activities without disrupting their natural task flow. This task can be formulated as a joint timing-and-action decision problem under the constraint that the human task remains the primary process.

To study this problem systematically, we establish the NIABench simulation benchmark (coupled with real-world experiments), built on the AI2-THOR simulator~\cite{kolve2017ai2} across four domestic environments (Kitchen, Bedroom, Living Room, Bathroom) with 118 interactive objects and 800 atomic actions. The benchmark includes 2K training episodes and seven representative test tasks, and is easily extensible for more episodes and diverse tasks; we also introduce three tailored metrics to rigorously evaluate model effectiveness for non-intrusive assistance.

Building on NIABench, we develop a dedicated non-intrusive assistance framework where robot behavior is constrained by both task utility and strict non-interruption rules—treating the human task plan as the primary, unalterable process and the robot as an auxiliary agent that decides when and what to act to reduce downstream human effort. This reframes the core challenge: the robot must infer well-timed, effective non-intrusive assistance without human negotiation or explicit queries. To solve this, we propose NiaRR (Non-intrusive Assistance with Retrieval and Ranking), a hybrid retrieval-and-ranking architecture balancing semantic generalization and fine-grained temporal reasoning: a semantic retriever~\cite{reimers2019sentence} first prunes the large action space to a compact relevant candidate set, and a transformer-based encoder then scores joint (step, action) pairs to identify the most beneficial non-intrusive assistance actions. This two-stage design ensures tractable inference while enabling rich cross-step reasoning in the ranking phase.

In summary, our contributions are as follows.
\begin{itemize}
  \item We formalize the non-intrusive assistance task in HRI, which focuses on supporting humans without disrupting their ongoing actions. To facilitate evaluation, we develop a corresponding simulation benchmark, NIABench, along with tailored metrics.
  \item We propose a simple yet effective approach that integrates semantic retrieval with joint scoring to efficiently and accurately identify non-intrusive assistance.
  \item We conduct comprehensive empirical evaluations on both NIABench and real-world settings, demonstrating that our method can substantially reduce human effort without intervention and improve decision quality without compromising task success.
\end{itemize}

\section{Related Work}
\label{sec:related_work}

\subsection{Human-Robot Interaction}

While early HRI work focused on human-robot coordination~\cite{goodrich2008human}, subsequent research advances along reactive, predictive, and proactive assistance lines. Reactive assistance relies on robots responding to explicit human instructions, with early studies on natural language grounding for navigation and manipulation~\cite{tellex2011, hemachandra2015learning, misra2016tell}, recent LLM-based zero-shot planning and adaptive instruction-following~\cite{huang2022language, luan2024enhancing, thumm2025text2interaction, grannen2025vocal}, and multimodal interfaces optimizing reactive control intuitiveness~\cite{campagna2025fostering, yu2025mixed}. 

Proactivity is a core direction for future HRI~\cite{li2023proactive}: predictive approaches enable intent inference and action prediction for human need anticipation~\cite{koppula2015anticipating, maeda2017phase, nikolaidis2017human, jain2018recursive, landi2019prediction, belardinelli2024gaze, hoffman2024inferring} (e.g., object affordances~\cite{koppula2015anticipating}, recursive Bayesian inference~\cite{jain2018recursive}) but are limited to short-term observed-behavior-based predictions; proactive assistance shifts initiative to robots via forgotten action reminders~\cite{wu2016watch}, gaze-based grasping support~\cite{shafti2019gaze}, spatio-temporal object modeling~\cite{patel2023proactive} and human-object interaction prediction~\cite{mascaro2023hoi4abot}, yet its direct human task interventions introduce extra cognitive load and resumption costs~\cite{lee2018interruption, hirsch2024examining}. 
Some studies in proactive assistance have explored non-intrusive design tendencies to reduce human cognitive load~\cite{schmid2007proactive, mollaret2016multi, harman2020action, quesada2022proactive}. 
However, these works still take the robot or environment as the assistance leader and allow robot to initiate adjustments to human task rhythms, ignoring the autonomy of the human task process. 

Different from prior interaction paradigms, we propose Non-intrusive Assistance, which treats the human plan as the primary and authoritative process. The robot acts opportunistically and silently, avoiding any need for human attention or replanning, in contrast to approaches that frame coordination as an explicit negotiation problem.

\subsection{Large Language Models in Robotics}

Non-intrusive assistance requires long-horizon reasoning for joint timing-and-action decisions with the human task as the primary process, where LLMs~\cite{achiam2023gpt, dubey2024llama} bridge high-level instructions and low-level robotic behaviors effectively~\cite{zeng2023large}. Early work~\cite{brohan2023can, huang2023inner, huang2022language, liu2023reflect} validated language grounding for robotic actions; follow-up studies explored LLMs as instruction-following planners~\cite{song2023llm, shin2024socratic}, integrated safety alignment into HRI~\cite{thumm2025text2interaction, wang2025apricot}, and fused multimodal perception for better grounding~\cite{ren2023robots, wang2024polaris, gao2024physically, murray2025teaching}. Yet LLM-based methods face real-time assistive challenges: overgeneralized outputs, irrelevant candidates, infeasible step hallucinations, and no explicit assistance timing reasoning, leaving silent, well-timed non-intrusive assistance unsolved. In this work, we use LLMs as a complementary component, leveraging their generative strengths and mitigating noise and timing limitations via explicit candidate pruning and joint scoring.

\section{Benchmark}
\label{sec:benchmark}

\begin{figure*}[t]
  \centering
  \includegraphics[width=1.0\textwidth]{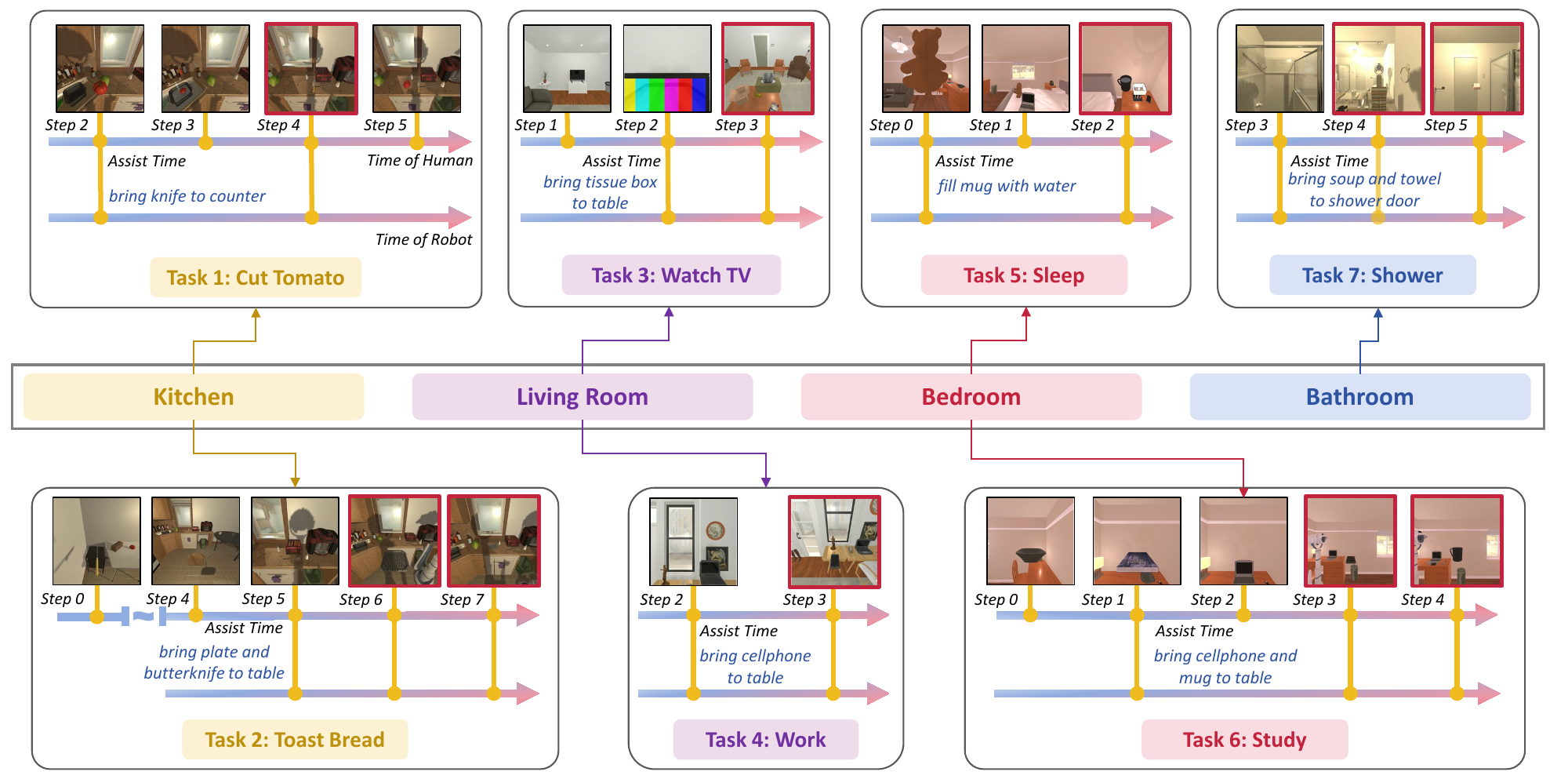}
  \caption{We visualize the seven evaluation episodes (tasks), which are similar in structure to the training episodes. Each episode contains information about two agents, representing the human and the robot, including the scene, the human task sequence, the robot assistance labels, and other relevant details.
  }
  \label{fig:benchmark}
\end{figure*}

We present NIABench based on AI2-THOR~\cite{kolve2017ai2}, a simulation benchmark with accompanying metrics designed to systematically evaluate non-intrusive assistance.
Since the simulator provides direct access to object and robot states, we simplify the setting by omitting raw visual perception and instead construct the entire benchmark in textual form. This design highlights the core challenge of non-intrusive assistance, namely the joint decision of timing and action, without being confounded by perception errors. Moreover, as modern vision models can also produce textual reasoning, they can be seamlessly integrated into the benchmark, further justifying this simplification. NIABench consists of (1) a corpus of scripted human-task episodes with oracle assistance task annotations, (2) a task suite spanning four domestic environments, and (3) a set of evaluation metrics that assess both per-decision quality and task-level human utility.

\subsection{Benchmark Construction}

\subsubsection{Scenes and Object Vocabulary}
NIABench is built on the AI2-THOR simulator~\cite{kolve2017ai2} and spans four indoor scene categories: \textit{Kitchen, Bedroom, Living Room, and Bathroom}. Each scene is equipped with a curated set of object classes and receptacles relevant to common household tasks (e.g., \texttt{Knife}, \texttt{Plate}, \texttt{CounterTop} in the Kitchen), totaling 118 interactive objects. To capture visual and spatial diversity, each object class includes multiple variants with different sizes, textures, and placements.

\subsubsection{Episode Format}
\label{subsubsec:episod_format}

We collected 2,000 training episodes in NIABench to train our model. Each episode (task) is represented as a deterministic and replayable demonstration, stored as a single record with the following fields:

\begin{itemize}
    \item \textbf{episode\_id}: A unique identifier for the episode. 
    \item \textbf{scene}: One of \textit{kitchen}, \textit{bedroom}, \textit{livingroom}, or \textit{bathroom}. 
    \item \textbf{human\_task\_seq}: A realistic sequence of atomic human actions to be executed (e.g., [\texttt{find\_tomato}, \allowbreak \texttt{bring\_tomato\_to\_sink}, \allowbreak \texttt{wash\_tomato}, \allowbreak \texttt{bring\_tomato\_to\_counter}, \allowbreak \texttt{bring\_knife\_to\_counter}, \allowbreak \texttt{cut\_tomato}]). This sequence constitutes the primary process that the robot must not interrupt. 
    \item \textbf{robot\_vocab}: The candidate set of high-level robot actions available for ranking within the episode. These actions are drawn from the scene-specific vocabulary (e.g., \texttt{find\_knife}, \texttt{clean\_countertop}, \texttt{bring\_knife\_to\_countertop}). 
    \item \textbf{oracle\_labels}: Annotation of each episode specifies a \texttt{human\_step\_idx}, indicating the human step at which the assistance task should occur, and a \texttt{best\_robot\_action} from \texttt{robot\_vocab}, denoting the desirable robot action at that step (e.g., [\{\texttt{human\_step\_idx}: 2, \texttt{best\_robot\_action}: \texttt{bring\_knife\_to\_countertop}\}]). 
\end{itemize}

\subsubsection{Annotation Distribution and Dataset Splits}

To capture realistic variability, episodes are first generated using an LLM~\cite{achiam2023gpt} conditioned on simulator scene information~\cite{kolve2017ai2}, and then manually checked and annotated by three trained HRI researchers (expert annotators). In the final dataset, 75\% of episodes contain exactly one oracle assistance task, 20\% contain two non-overlapping assistance tasks, and 5\% contain no assistance task, where each episode was annotated by two independent annotators. We further split the corpus into training and validation folds, stratified by scene and by the number of oracle labels per episode, to preserve the distribution of assistance task counts across splits.

\subsubsection{Evaluation in Simulation}
We additionally design a held-out suite of seven representative simulator tasks, distinct from the training set, which serve as test tasks to evaluate model performance, as shown in Figure~\ref{fig:benchmark}. Detailed descriptions of these episodes are provided in the appendix of the supplementary materials.

\subsection{Evaluation Metrics}

Our benchmark evaluates the performance of methods on non-intrusive assistance at two levels: decision-level metrics, which assess how accurately the model selects oracle pairs, and task-level utility metrics, which measure the impact of assistance when the model is executed in NIABench.

\subsubsection{Decision-level Metrics}
We propose \textbf{SelectionAcc}, defined as the fraction of cases where the model’s top-1 predicted action matches the oracle assistance task label. For episodes with multiple oracle annotations, each human-step and oracle-action pair is evaluated independently.

\subsubsection{Task-level Utility Metrics}
To evaluate task-level performance, we run each test episode in NIABench with the trained model controlling the robot while the human executes the scripted sequence. Let $H_{\mathrm{human}}$ denote the number of simulator-executed atomic steps in the human-only baseline, and $H_{\mathrm{assist}}$ the number of simulator-executed when assisted by the robot. We define \textbf{HumanStepSaved (HSS)} as $\mathrm{HSS} = H_{\mathrm{human}} - H_{\mathrm{assist}}$ and report the mean absolute HSS over test episodes, which directly measures how many steps of human effort are reduced through assistance.

We also report \textbf{SuccessAcc}, defined as the fraction of evaluation episodes in which the overall task is successfully completed at the end of the assisted run. Task success is computed deterministically from the simulator state, taking into account both object positions and object state attributes.

\begin{figure*}[t]
  \centering
  \includegraphics[width=0.95\textwidth]{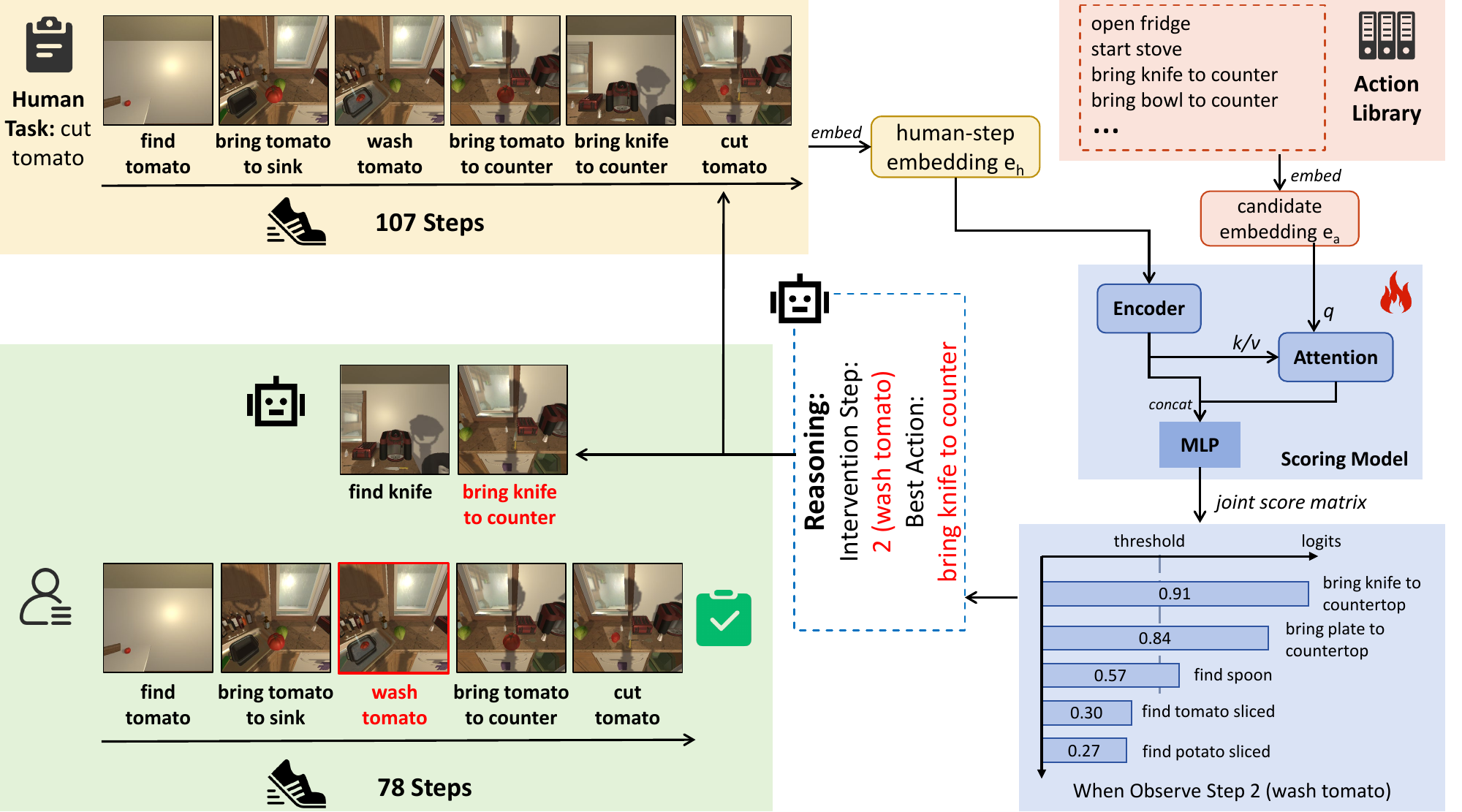}
  \caption{
  Overview of our framework for non-intrusive assistance. Given a human task sequence and candidate robot actions from the action library, both are first encoded with a pretrained encoder~\cite{reimers2019sentence}. The resulting embeddings are then processed by a trainable scoring model, which applies cross-step attention to produce a joint step–action score matrix. The top-scoring non-intrusive robot action is then selected and executed. 
  }
  \label{fig:model}
\end{figure*}

\section{Method}
\label{sec:method}

In this section, we present our approach to achieving non-intrusive human–robot interaction. 
We first define the problem setting, where a human task consists of sequential steps and the robot possesses a library of candidate assistive actions. 
We then introduce our method NiaRR, illustrated in Figure~\ref{fig:model}, which aligns human task steps with suitable assistive actions by leveraging token embeddings and transformer-based attention mechanisms.
Finally, we describe the training objectives and inference procedure that ensure the selected actions remain supportive yet non-intrusive.

\subsection{Problem Formulation}
We consider non-intrusive human–robot assistance in multi-step scripted tasks. 
A human task is represented as a sequence 
$\mathcal{H} = (h_1,\dots,h_S)$, 
where $h_s$ denotes the $s$-th atomic human step (e.g., \texttt{find\_apple}, \texttt{wash\_apple}), with $s \in \{1,\dots,S\}$ and $S$ the sequence length where each atomic step maps to a set of simulator-executed primitive steps. 
For a given scene, the robot has a context-dependent library of high-level executable actions 
$\mathcal{A}=\{a_0,\dots,a_C\}$, where $a_0=\texttt{no\_op}$ denotes a special ``no operation'' action indicating that the robot remains idle and offers no assistance. Thus, $|\mathcal{A}|=C{+}1$. 
Given a human sequence $\mathcal{H}$ and an action library $\mathcal{A}$, the task is to identify a set of assistive pairs $(s,a)$, where each pair specifies that the robot should execute action $a$ at step $s$ provided the action supports the task without interrupting the ongoing process.
Ground-truth labels $(s^\star,a^\star)$ are available during training, indicating desirable human–robot alignments, while at test time the goal is to predict high-quality assistive actions consistent with the non-intrusiveness requirement.

\subsection{Representation and Candidate Retrieval}
\label{sec:repre}

We obtain vector embeddings for all tokens using a Sentence-BERT (SBERT) encoder \cite{reimers2019sentence, wang2020minilm}:
\begin{equation}
e_h(s) = \mathrm{emb}(h_s), \quad e_a(c) = \mathrm{emb}(a_c),
\end{equation}
where $e_h(s), e_a(c) \in \mathbb{R}^{D}$, and $D$ is the embedding dimension of the SBERT model. 
For an episode with human sequence length $S$ and candidate set size $C$, we construct
$H_{\text{emb}} \in \mathbb{R}^{S\times D} = [e_h(1), \dots, e_h(S)]^\top$ and
$A_{\text{emb}} \in \mathbb{R}^{C\times D} = [e_a(1), \dots, e_a(C)]^\top$.
To improve efficiency, we prune the candidate set by cosine similarity:
\begin{equation}
    \mathrm{cos}(h_s,a_c) = \frac{e_h(s)\cdot e_a(c)}{\|e_h(s)\|\|e_a(c)\|},
\end{equation}
keeping the top-$K$ actions per step. 
This yields a compact candidate set $\mathcal{A}'_s$ for subsequent ranking.

\subsection{Joint Step -- Action Scoring}
\label{sec:scoring}

We estimate the compatibility between human steps and robot candidate actions using a Transformer with cross-attention. 
Let $\mathbf{H}_{\text{emb}}\in\mathbb{R}^{B\times S\times D}$ and 
$\mathbf{A}_{\text{emb}}\in\mathbb{R}^{B\times C\times D}$ denote SBERT embeddings of human steps and candidate actions, 
where $B$ is the batch size.
Both are projected to a common model dimension $D_m$:
\begin{equation}
\widetilde{\mathbf{H}}=\operatorname{proj}_h(\mathbf{H}_{\text{emb}}), 
\qquad 
\widetilde{\mathbf{A}}=\operatorname{proj}_a(\mathbf{A}_{\text{emb}}).
\end{equation}

\noindent \textbf{Step Encoding.}  
We add sinusoidal positional encodings to $\widetilde{\mathbf{H}}$ to preserve step order, 
and then apply a Transformer encoder with an attention mask that excludes padded steps: 
\begin{equation}
\mathbf{H}=\operatorname{Encoder}(\widetilde{\mathbf{H}}+\mathrm{PE}(1{:}S))
   \in\mathbb{R}^{B\times S\times D_m}.
\end{equation}

\noindent \textbf{Cross-attention.}  
Candidate embeddings $\widetilde{\mathbf{A}}$ serve as queries attending over $\mathbf{H}$:
\begin{equation}
\mathrm{Attn}(Q,K,V) = \mathrm{softmax}\!\left(\tfrac{QK^\top}{\sqrt{d_k}}\right)V,
\end{equation}
where $d_k=D_m/h$ the per-head key dimension for $h$ attention heads and $Q=\widetilde{\mathbf{A}}, K=V=\mathbf{H}$.
This produces action-aware representations $\mathbf{A}^{\mathrm{att}}\in\mathbb{R}^{B\times C\times D_m}$.

\noindent \textbf{Pair Scoring.}  
For each step $s$ and candidate $c$, we form
\begin{equation}
\mathbf{p}_{s,c} = [\,\mathbf{H}_{s,:}\;\|\;\mathbf{A}^{\mathrm{att}}_{c,:}\,]
   \in\mathbb{R}^{2D_m},
\end{equation}
and score it with an MLP. Collecting all scores yields
\begin{equation}
\mathrm{Logits}\in\mathbb{R}^{B\times S\times C},
\end{equation}
where $\mathrm{Logits}_{b,s,c}$ is the score of candidate $c$ at step $s$ in example $b$, 
and invalid steps/candidates are masked before softmax.

\subsection{Traing and Inference}

\noindent \textbf{Training.}  
Given oracle labels $(s^\star,a^\star)$, we flatten the logits tensor into 
$\mathbb{R}^{B\times (S\cdot C)}$ and optimize a cross-entropy loss:
\begin{equation}
\mathcal{L}_{\text{CE}} = -\tfrac{1}{B}\sum_{b=1}^B 
\log \frac{\exp(\mathrm{Logits}^{\text{flat}}_{b,y_b})}
{\sum_{j=1}^{S\cdot C}\exp(\mathrm{Logits}^{\text{flat}}_{b,j})},
\end{equation}
where $y_b$ is the flattened index corresponding to the oracle step--action pair for example $b$.  

\noindent \textbf{Inference.}  
At test time, predictions are obtained by selecting the highest-scoring pair:
\begin{equation}
(\hat s,\hat c)=\arg\max_{s,c}\;\mathrm{Logits}_{s,c},
\end{equation}
which yields the assistive action $\hat a=a_{\hat c}$, including $\texttt{no\_op}$ as a valid option.

\noindent \textbf{Operation Details.}
Since the simulator only supports a small set of primitive API calls 
(e.g., \texttt{MoveTo}, \texttt{PickupObject}, \texttt{PutObject}), 
we employ a lightweight adapter that translates predicted high-level tokens 
(e.g., \texttt{bring\_knife\_to\_countertop}) into such primitives. 
This component is purely operational and not part of the learning model.

\section{Experiments}
\label{sec:exp}

\subsection{Experimental Setup}

\noindent \textbf{Benchmark Tasks.} 
We evaluate on NIABench, a benchmark for non-intrusive assistance introduced in Section~\ref{sec:benchmark}.
It covers four scene types: \textit{kitchen}, \textit{bedroom}, \textit{living\_room}, and \textit{bathroom}, 
and includes 2,000 training tasks and seven representative evaluation tasks: \texttt{cut\_tomato}, \texttt{toast\_bread}, \texttt{watch\_tv}, 
\texttt{work}, \texttt{sleep}, \texttt{study}, and \texttt{shower}.
We also evaluate our method on two real-world tasks: assisting with peeling an apple and cleaning a table.

\noindent \textbf{Baselines.}
We evaluated our method against different baselines including GPT4 only~\cite{achiam2023gpt}, GPT4 with zero-shot chain-of-thought (CoT) prompting ~\cite{zhang2022automatic}, GPT4 with few-shot CoT~\cite{zhang2022automatic}, ReAct~\cite{yao2023react}, and RE2~\cite{xu2024re}.
To ensure fair comparison, all baselines share identical data preprocessing and candidate vocabulary choices, and the only differences lie in the model architecture or the sourcing of candidate actions.

\noindent \textbf{Metrics.} 
For each task, we evaluate performance using three metrics designed to capture non-intrusive assistance: HumanStepSaved (HSS), SuccessAcc, and SelectionAcc, with definitions provided in Section~\ref{sec:benchmark}.

\noindent \textbf{Implementation details.} 
We use SBERT~\cite{reimers2019sentence}, with embeddings for all action tokens pre-computed and cached. The transformer backbone is configured with a model dimension of $D_m=256$, $L=2$ encoder layers, and $H=4$ attention heads. Training is performed with AdamW using a learning rate of 3e-4, a weight decay of 1e-2, for 12 epochs with a batch size of 64. For efficiency, the robot vocabulary is restricted to the top-20 actions retrieved from the action library. During execution, the robot follows a non-blocking protocol to ensure that ongoing human tasks are never interrupted, 

which avoids the robot preempting human resources or spaces and performs auxiliary actions in parallel or between human steps.

\begin{table*}[t]
\centering
\caption{Results of all methods on seven NIABench tasks. For each task, we report HumanStepSaved (HSS) and SuccessAcc (\%) metrics.}
\label{tab:main_exp}
\begin{tabular}{l cc cc cc cc}
\toprule
 & \multicolumn{2}{c}{\makecell{Kitchen\\cut tomato}} 
& \multicolumn{2}{c}{\makecell{Kitchen\\toast bread}} 
& \multicolumn{2}{c}{\makecell{Livingroom\\watch tv}} 
& \multicolumn{2}{c}{\makecell{Livingroom\\work}} \\
\cmidrule(lr){2-3}\cmidrule(lr){4-5}\cmidrule(lr){6-7}\cmidrule(lr){8-9}
Method & HSS $\uparrow$ & SuccessAcc $\uparrow$ & HSS $\uparrow$ & SuccessAcc $\uparrow$& HSS  $\uparrow$& SuccessAcc $\uparrow$ & HSS $\uparrow$& SuccessAcc $\uparrow$ \\
\midrule
GPT4~\cite{achiam2023gpt} & 7.0 & 94.0 & 9.0 & 86.0 & 6.0 & 93.0 & 5.0 & 89.0 \\
GPT4 (with zero-shot CoT)~\cite{zhang2022automatic} & 9.0 & 95.0 & 16.0 & 86.0 & 6.0 & 95.0 & 15.0 & 91.0 \\
GPT4 (with few-shot CoT)~\cite{zhang2022automatic} & 11.0 & 95.0 & 19.0 & 89.0 & 7.0 & 95.0 & 16.0 & 91.0 \\
ReAct~\cite{yao2023react} & 13.0 & 96.0 & 28.0 & 89.0 & 7.0 & 95.0 & 24.0 & 93.0 \\
RE2~\cite{xu2024re} & 17.0 & \textbf{99.0} & 31.0 & 92.0 & 7.0 & 96.0 & 26.0 & \textbf{95.0} \\
\rowcolor{gray!9}\textbf{NiaRR (Ours)} & \textbf{29.0} & \textbf{99.0} & \textbf{38.0} & \textbf{95.0} & \textbf{16.0} & \textbf{98.0} & \textbf{37.0} & \textbf{95.0} \\
\bottomrule
\end{tabular}
\vspace{1ex}

\begin{tabular}{l cc cc cc cc}
\toprule
& \multicolumn{2}{c}{\makecell{Bedroom\\sleep}} 
& \multicolumn{2}{c}{\makecell{Bedroom\\study}} 
& \multicolumn{2}{c}{\makecell{Bathroom\\shower}} & \multicolumn{2}{c}{Avg.} \\
\cmidrule(lr){2-3}\cmidrule(lr){4-5}\cmidrule(lr){6-7}\cmidrule(lr){8-9}
Method & HSS $\uparrow$ & SuccessAcc $\uparrow$ & HSS $\uparrow$ & SuccessAcc $\uparrow$ & HSS $\uparrow$ & SuccessAcc $\uparrow$ & HSS $\uparrow$& SuccessAcc $\uparrow$ \\
\midrule
GPT4~\cite{achiam2023gpt} & 3.0 & 71.0 & 18.0 & 74.0 & 21.0 & 75.0 & 9.9 & 83.1 \\
GPT4 (with zero-shot CoT)~\cite{zhang2022automatic} & 3.0 & 77.0 & 22.0 & 80.0 & 26.0 & 79.0 & 13.9 & 86.1 \\
GPT4 (with few-shot CoT)~\cite{zhang2022automatic} & 3.0 & 80.0 & 25.0 & 79.0 & 26.0 & 81.0 & 15.3 & 87.1 \\
ReAct~\cite{yao2023react} & 4.0 & 84.0 & 26.0 & 80.0 & 27.0 & 85.0 & 18.4 & 88.9 \\
RE2~\cite{xu2024re} & 4.0 & 87.0 & 26.0 & 84.0 & 30.0 & 89.0 & 20.1 & 91.7 \\
\rowcolor{gray!9}\textbf{NiaRR (Ours)} & \textbf{13.0} & \textbf{93.0} & \textbf{35.0} & \textbf{89.0} & \textbf{38.0} & \textbf{96.0} & \textbf{29.4} & \textbf{95.0} \\
\bottomrule
\end{tabular}

\end{table*}

\begin{table*}[t]
\centering
\caption{Analysis of SelectionAcc (\%) metric results on seven NIABench tasks.}
\label{tab:analysis}
\begin{tabular}{l c c c c c c c c}
\toprule

Method & \makecell{Kitchen\\cut tomato} & \makecell{Kitchen\\toast bread} & \makecell{Livingroom\\watch tv} & \makecell{Livingroom\\work} & \makecell{Bedroom\\sleep} & \makecell{Bedroom\\study} & \makecell{Bathroom\\shower} & Avg. \\
\midrule
GPT4 only~\cite{achiam2023gpt} & 61.0 & 53.0 & 32.0 & 21.0 & 37.0 & 38.0 & 71.0 & 44.7 \\
GPT4 (with zero-shot CoT)~\cite{zhang2022automatic} & 65.0 & 61.0 & 33.0 & 30.0 & 39.0 & 41.0 & 77.0 & 49.4 \\
GPT4 (with few-shot CoT)~\cite{zhang2022automatic} & 67.0 & 66.0 & 37.0 & 33.0 & 39.0 & 44.0 & 80.0 & 52.3 \\
ReAct~\cite{yao2023react} & 73.0 & 68.0 & 50.0 & 41.0 & 48.0 & 55.0 & 80.0 & 59.3 \\
RE2~\cite{xu2024re} & 78.0 & 73.0 & 54.0 & 44.0 & 52.0 & 57.0 & 84.0 & 63.1 \\
\rowcolor{gray!9}\textbf{NiaRR (Ours)} & \textbf{96.0} & \textbf{86.0} & \textbf{73.0} & \textbf{67.0} & \textbf{77.0} & \textbf{76.0} & \textbf{94.0} & \textbf{81.3} \\
\bottomrule
\end{tabular}
\end{table*}

\begin{table}[t]
\centering
\caption{Ablation results on cut\_tomato task.}
\label{tab:ablation}
\begin{tabular}{lcc}
\toprule
Setting & HumanStepSaved & SuccessAcc (\%) \\
\midrule
\rowcolor{gray!9} \textbf{NiaRR (Ours)} & \textbf{29.0} & \textbf{99.0} \\
~~ w/o top-$K$ retrieval & 23.0 & 91.0 \\
~~ w/o joint S$\times$C & 19.0 & 93.0 \\
\bottomrule
\end{tabular}
\end{table}

\subsection{Results on NIABench}
We report the performance of NIABench across various methods and our approach in Table~\ref{tab:main_exp}. Our method consistently outperforms all baselines, achieving the highest HSS on every task and the highest or near-highest SuccessAcc. On average, it saves 29.4 human steps and attains a SuccessAcc of 95.0\%, both the best among all methods. These results demonstrate that the proposed approach improves human effort reduction while maintaining or improving task reliability, thereby providing effective non-intrusive assistance.

\subsection{Ablation and Analysis Study}

\subsubsection{Ablation Study}

We performed two ablations on the \texttt{cut\_tomato} task. 
First, w/o joint $S\times C$ means that the model is forced to predict only the robot action without the relevant human step, and in inference, we use a simple attention-based approach to pick a step from the human task sequence. 
Second, w/o top-$K$ retrieval means that we remove the SBERT-based $top-K$ semantic pre-screening so that the transformer ranker must choose from the full candidate list of each episode data during both training and inference. 

The results are as shown in Table~\ref{tab:ablation}, which demonstrate that: 
First, removing the joint $S \times C$ component and adopting the action-only variant weakens the model's reasoning ability, and even if the optimal action is selected, the execution timing remains more suboptimal, making the robot unable to effectively reduce the same number of human steps. This indicates that when to act and what to do is important to generate truly useful and non-intrusive assistance. 
Second, removing top-$K$ retrieval usually reduces HumanStepSaved and SuccessAcc metrics, because the lack of semantic retrieval exposes the model to more noisy and irrelevant candidates, which weakens the model's selection accuracy. 
In summary, the ablation experiments validate the essential of semantic candidate retrieval and joint step--action ranking for non-intrusive assistance.

\subsubsection{Analysis on Selection Accuracy}

We analyze SelectionAcc, which measures whether the model’s inferred action matches the oracle label for non-intrusive robot assistance, across seven NIABench tasks. As shown in Table~\ref{tab:analysis}, our method achieves the highest selection accuracy compared to other LLM-based baselines on all tasks, demonstrating the effectiveness of our method. 
Moreover, these results complement the primary task-level metrics, HSS and SuccessAcc, indicating that higher SelectionAcc correlates with better human effort reduction and more reliable task completion.  

\begin{figure*}[t]
  \centering
  \includegraphics[width=0.8\textwidth]{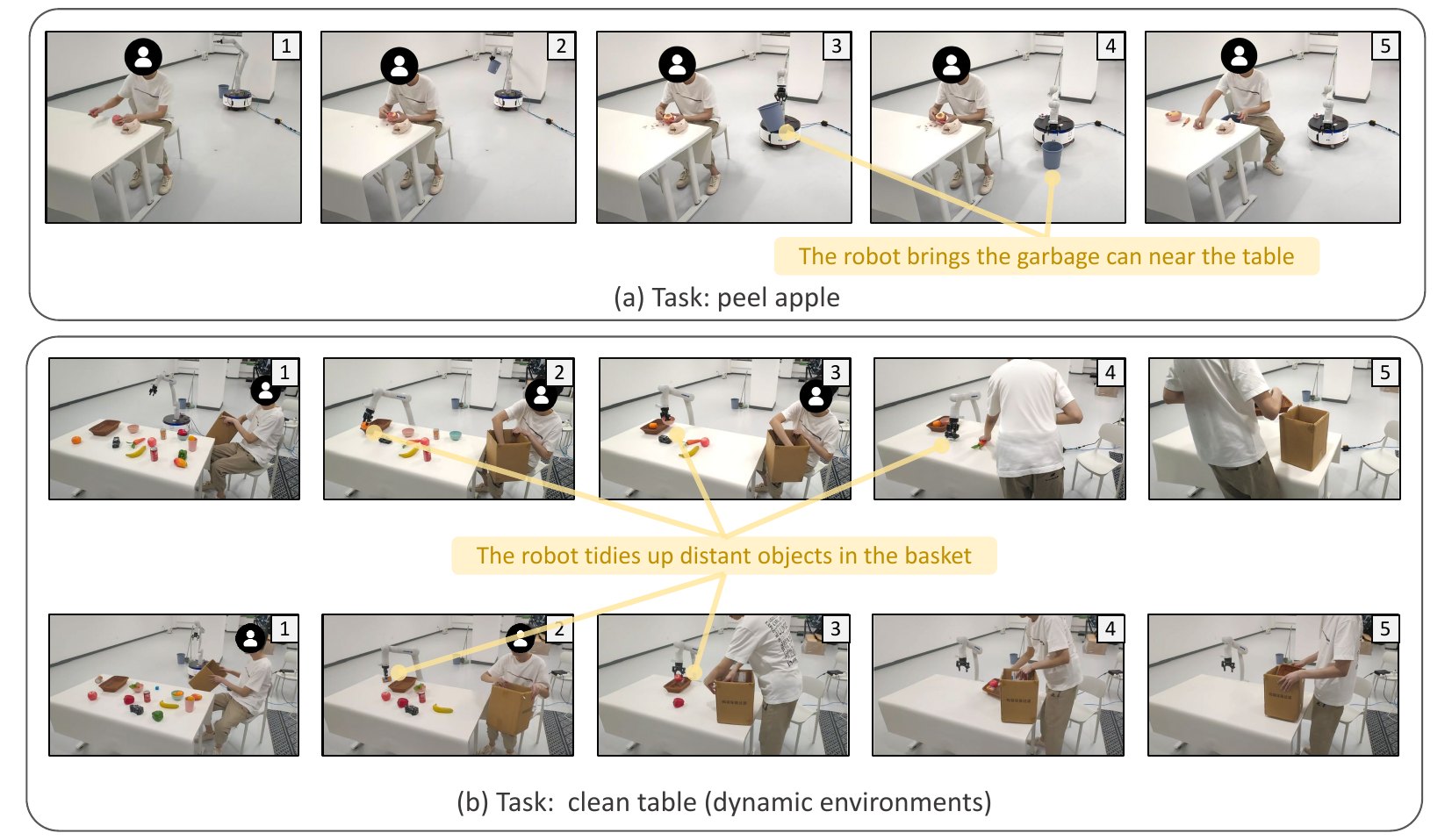}
  \caption{Real-robot demonstrations of two tasks. (a) The robot assists a human in peeling an apple. (b) The robot assists a human in cleaning a table in a dynamic environment.}
  \label{fig:real_robot}
\end{figure*}

\subsection{Real-World Experiments}
To demonstrate the feasibility of non-intrusive assistance on a physical robot, we conducted two non-scripted real-world tasks beyond NIABench. These tasks leveraged a navigation algorithm for the mobile base, a vision–language model for object recognition~\cite{team2025robobrain}, and a grasping model~\cite{zhang2025vcot} for object manipulation. In the \texttt{peel\_apple} task (Figure~\ref{fig:real_robot} (a), the human sat at a table slicing an apple while the robot opportunistically fetched and positioned a garbage can nearby, eliminating the need for the human to search for it to dispose of peelings. The assistive action was carried out without interrupting the user and effectively reduced the subsequent human steps. In the \texttt{clean\_table} task (Figure~\ref{fig:real_robot} (b), the human placed objects into a box while the robot simultaneously moved more distant items into a basket. We evaluated two layout variants with substantially different object placements and distances. In both cases, the robot successfully executed the assistance task, demonstrating dynamic adaptability in deciding which objects to move and when to act based on the current spatial configuration and available timing windows.

\section{Conclusion and Limitations}

We formalize and advance the non-intrusive assistance paradigm in Human–Robot Interaction (HRI), a well-studied direction in proactive assistive robotics. Unlike reactive assistance driven by explicit commands or proactive assistance with direct interventions, this paradigm emphasizes effective human support without disrupting ongoing task progress. To enable its systematic evaluation, we develop the NIABench simulation benchmark with tailored quantification metrics, and design a hybrid model fusing LLM-based semantic retrieval and a scoring module to prune irrelevant candidate actions and avoid hallucinated infeasible steps. 
We validate the effectiveness of our approach through experiments on both NIABench and real robots, demonstrating its ability to reduce human effort while maintaining task reliability.

Despite these contributions, our work has key limitations. First, non-disruptive human need inference requires deep insights into human habits and task structures, rendering the problem inherently challenging and necessitating simplifications in task design. Second, NIABench itself does not incorporate visual modalities due to the substantial simulation-reality gap, even as our real-robot experiments leverage pretrained vision-language and robotic manipulation models. In future work, we envision integrating multi-modal perception into non-intrusive assistive HRI to achieve more capable and realistic robotic assistance, and further address dynamic human plan changes and partial observability in real-world scenarios.

\bibliography{main, IEEEabrv}

\end{document}